\documentclass[journal]{IEEEtran} % two column single spacing
\usepackage{array}
\usepackage{graphicx}
\usepackage{amsmath,amssymb,amsthm, amsfonts}
\usepackage{resizegather}
\usepackage{tabularx} % for the table
\usepackage[utf8]{inputenc}
\usepackage[english]{babel}
\DeclareMathOperator{\MSE}{MSE}  % mean squared error
\newcommand{\density}{\ensuremath{{D}}} % density or p of the Bernoulli distribution
\newcommand{\p}{\density} % legacy ...
\newcommand{\numSamples}{\ensuremath{N}} % density or p of the Bernoulli distribution
\newcommand{\N}{\numSamples} % legacy ...

\newcommand{\occluder}{\ensuremath{O}}

\newcommand{\visibility}{\ensuremath{V}}

\newcommand{\img}{\ensuremath{S}} 
\newcommand{\R}{\img}  % for Bettina's equations
\newcommand{\noise}{\occluder}

\newcommand{\X}{\ensuremath{X}}  % integral/average image
\newcommand{\E}{\operatorname{E}}
\newcommand{\Ex}[1]{\ensuremath{\E[#1]}}

\newcommand{\sigmaOccl}{\ensuremath{\sigma_{o}^2}} 
\newcommand{\sigmaSign}{\ensuremath{\sigma_{s}^2}} 
\newcommand{\meanOccl}{\ensuremath{\mu_{o}}} 
\newcommand{\meanSign}{\ensuremath{\mu_{s}}} 

\newcommand{\appendixMath}[1]{\newline{}\scalebox{.8}{\parbox{1.25\linewidth}{#1}}\newline{}}
\begin{document}

\title{Fast Automatic Visibility Optimization \\ for Thermal Synthetic Aperture Visualization}

\author{Indrajit~Kurmi,
        David~C.~Schedl,
        and~Oliver~Bimber,~\IEEEmembership{Senior~Member,~IEEE}% <-this % stops a space
\thanks{I. Kurmi, D. Schedl, and O. Bimber are with the Computer Science Department of the Johannes Kepler University, Linz, Austria, e-mail: firstname.lastname@jku.at}% <-this % stops a space

\thanks{Manuscript accepted April 08, 2020}\vspace{-.1em}}

%\markboth{{This paper appears in:} IEEE GEOSCIENCE AND REMOTE SENSING LETTERS, 2020, DOI: {10.1109/LGRS.2020.2987471}}%
%{Kurmi \MakeLowercase{\textit{et al.}}: Airborne Optical Sectioning for Search and Rescue}
% *** FOR arXiv ***
\markboth{DOI {10.1109/LGRS.2020.2987471}, IEEE G\MakeLowercase{EOSCIENCE AND} R\MakeLowercase{EMOTE} S\MakeLowercase{ENSING} L\MakeLowercase{ETTERS}. \copyright~2020 IEEE. P\MakeLowercase{ersonal use only.}}{Kurmi \MakeLowercase{\textit{et al.}}: Airborne Optical Sectioning for Search and Rescue}

\maketitle

%------------------------------------------------------------------
\begin{abstract}
In this article, we describe and validate the first fully automatic parameter optimization for thermal synthetic aperture visualization. It replaces previous manual exploration of the parameter space, which is time consuming and error prone. We prove that the visibility of targets in thermal integral images is proportional to the variance of the targets' image. Since this is invariant to occlusion it represents a suitable objective function for optimization. Our findings have the potential to enable fully autonomous search and recuse operations with camera drones.
\end{abstract}
%------------------------------------------------------------------
\begin{IEEEkeywords}
Image Processing and Computer Vision, Enhancement.
% possible keywords from https://ieeecs-media.computer.org/assets/pdf/taxonomy.pdf
%I.2.9 Robotics
%I.2.9.a Autonomous vehicles
%I.2.9.l Vision
%I.2.10 Vision and Scene Understanding
%I.4 Image Processing and Computer Vision
%I.4.3 Enhancement
%I.5.4.b Computer vision
%G.1.6 Optimization
%G.1.6.a Constrained optimization
%
\end{IEEEkeywords}
\vspace{-0.5em}
% list of keywords: https://ieeecs-media.computer.org/assets/pdf/taxonomy.pdf
%------------------------------------------------------------------
\IEEEpeerreviewmaketitle
%------------------------------------------------------------------
\section{Introduction}
\IEEEPARstart{S}{earch} and rescue (SAR) operations often require to find lost or accidented people within densely forested terrain. While sunlight is mostly blocked by trees and other vegetation, little is reflected from forest ground. Therefore, thermal imaging is applied for measuring the temperature difference between human bodies and the surrounding environment. Yet, strong occlusion makes thermal imaging challenging. 

Synthetic apertures (SA) sensing samples the signal of wide aperture sensors with either arrays of static or single moving smaller aperture sensors whose individual signals are computationally combined to increase resolution, depth-of-field, frame rate, contrast, and signal-to-noise ratio. This principle has been used for radar, telescopes, microscopes, sonar, ultrasound, laser, and optical imaging \cite{Bimber2019IEEECGA}. With Airborne Optical Sectioning (AOS) \cite{Kurmi2019Sensor,Kurmi2018,Kurmi2019TAOS}, we have introduced a synthetic aperture imaging technique that captures an unstructured light field with an aircraft. Color and thermal images recorded within the shape of a wide (possibly hundreds to thousands of square meters) synthetic aperture area above forest are combined computationally to remove occluders, such as trees and other vegetation. The outcome is a widely occlusion free view of the forest ground. We implement AOS with autonomous camera drones, as they are becoming more and more relevant to SAR. They offer higher flexibility at lower cost compared to manned helicopters. However, AOS can be applied to any manned or unmanned aerial vehicle.    

Synthetic aperture visualization (i.e., the computational integration of individual recordings \cite{Kurmi2018}) requires unknown parameters for optimal occlusion removal that are, thus far, found interactively. A manual exploration of the parameter space (i.e., by visually evaluating interactive visualization results), however, is time consuming and error prone. In this article, we present a first, fast, and automatic parameter optimization for thermal synthetic aperture visualization.       
%------------------------------------------------------------------
%---------------------------------------------------
\begin{figure}[t!]
    \centering
    \includegraphics[width=.9\linewidth]{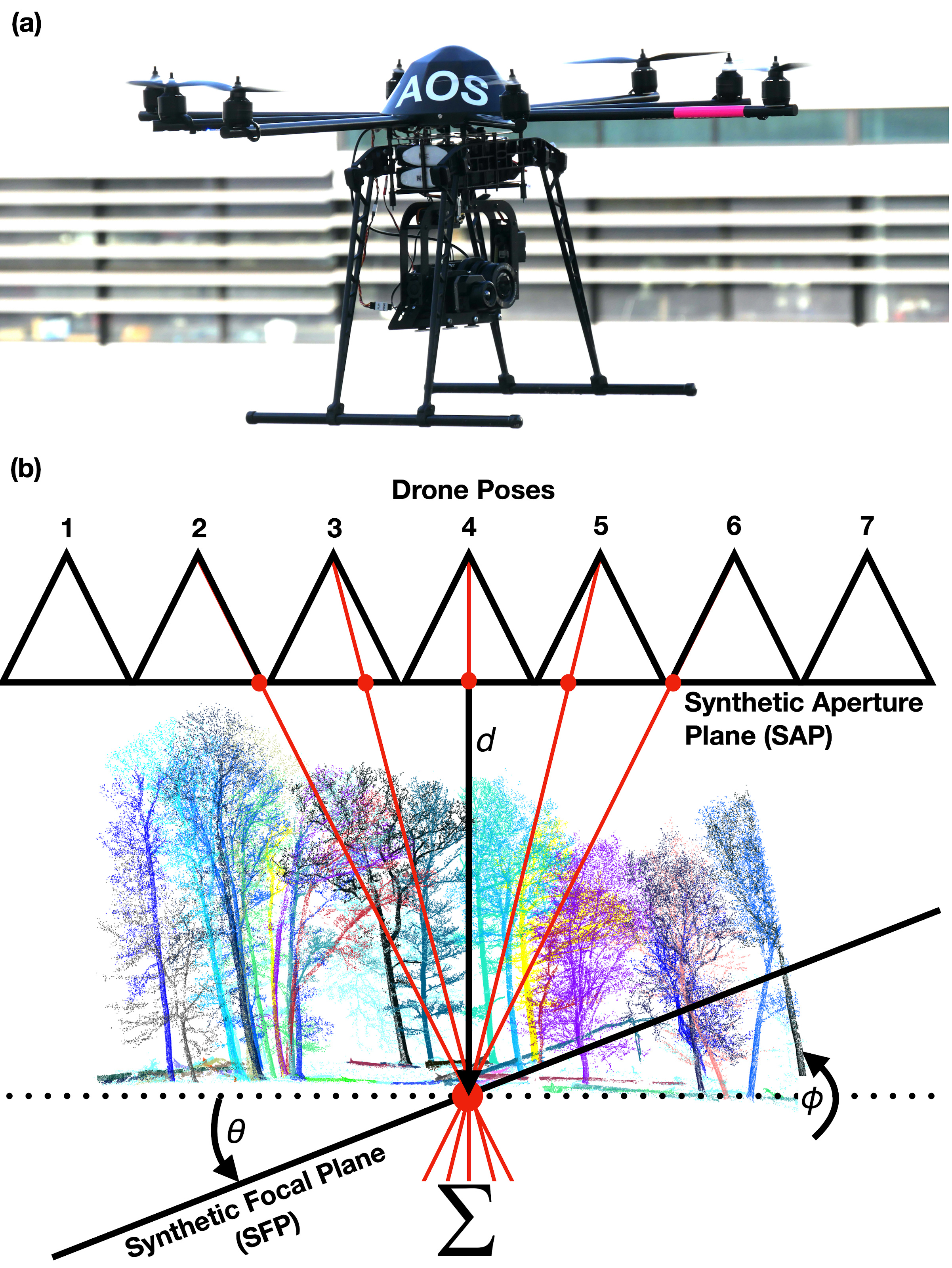}
    \caption{(a) Our drone equipped with synchronized RGB and thermal cameras on a rotatable gimbal. (b) The principle of synthetic aperture visualization, including the synthetic focal plane (SFP) parameters defined relative to the synthetic aperture plane (SAP): the distance $d$, and the tilt orientation angles $\theta,\phi$ in polar coordinates.}
    \label{fig:Fig1}
    \vspace{-1.0em}
\end{figure}
%---------------------------------------------------
%---------------------------------------------------
\begin{figure*}[t!]
    \centering
    \includegraphics[width=\linewidth]{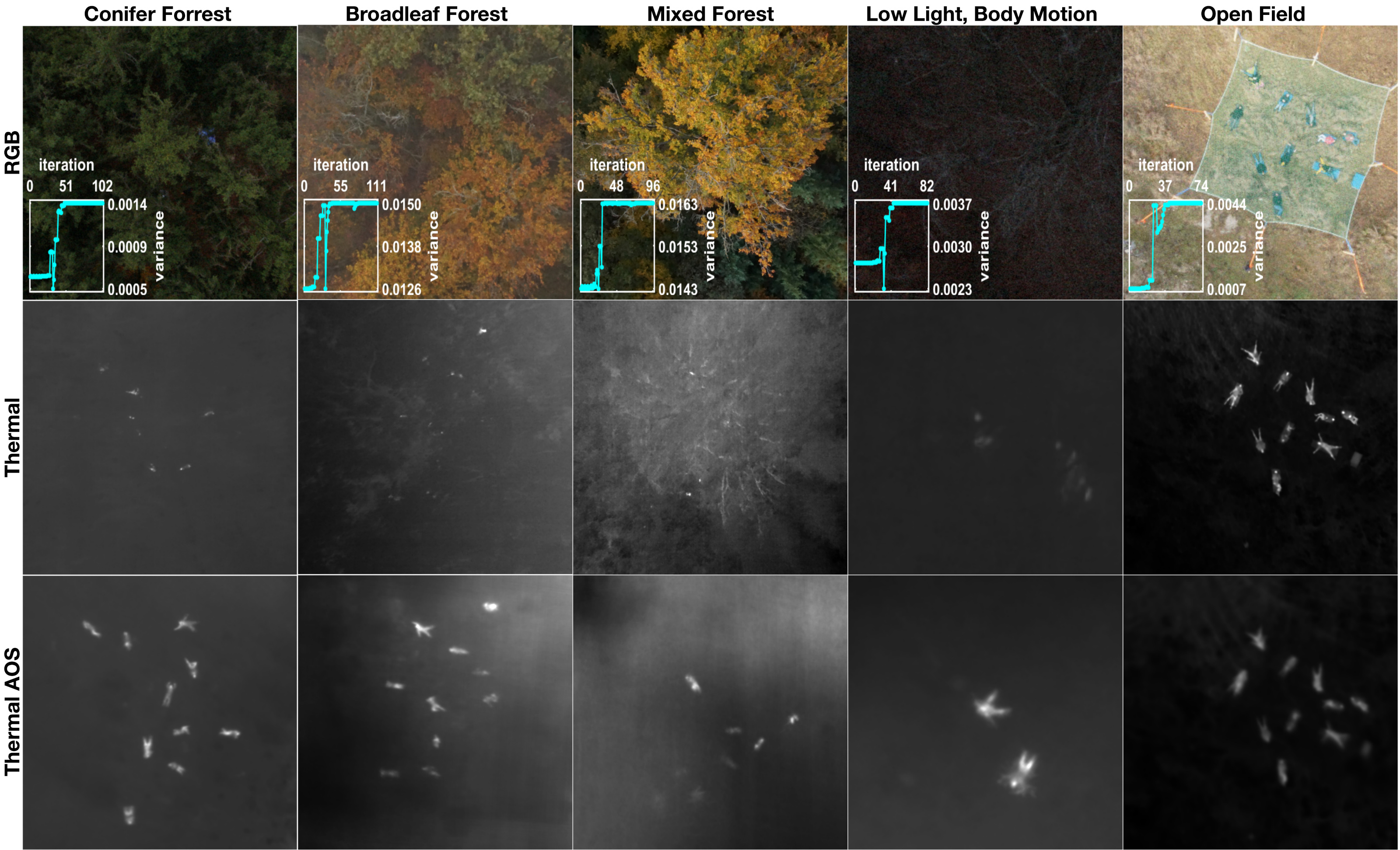}
    \caption{AOS scans of various environments (conifer forest, broadleaf forest, mixed forest, and open field), different lighting conditions (day light and low light), as well as for static and moving targets. The single RGB (upper row) and thermal (center row) recordings show the high degree of occlusion. The thermal integral images (bottom row) have been computed with the presented automatic optimization. All images in each columns depict the same view. The plots in the top row indicate the determined variance per iteration during local optimization (sequential quadratic programming). The final iteration found the global optimum in all cases.
    The scenes were sampled at 30-35 \textit{m} AGL, within an SA area of 900 \textit{$m^2$}, and with 340 recordings for the the conifer, broadleaf, mixed forest, and open field scenes; SA area was 79 \textit{$m^2$} with 31 recordings for the low light, body motion scene.}
    \label{fig:Fig3}
    \vspace{-1.0em}
\end{figure*}
%---------------------------------------------------
\section{Synthetic Aperture Imaging and Visualization}
As shown in Fig. \ref{fig:Fig1}a, we apply a redundant MikroKopter OktoXL 6S12 octocopter with a Flir Vue Pro thermal camera (9 $mm$ fixed focal length lens,  14 $bit$ tonal range covering a 7.5-13.5 $\mu m$ spectral band) and a Sony Alpha 6000 RGB camera (16-50 $mm$ lens, set to infinite focus). The drone autonomously records RGB-thermal image pairs at predefined positions with a chosen synthetic aperture area above forest. After recording, each image pair is rectified and the set of rectified RGB images are used for pose estimation using general-purpose structure-from-motion and multi-view stereo \cite{schonberger2016sfm,schoenberger2016mvs}). Intrinsic parameters and extrinsic transformation between RGB and thermal cameras are pre-calibrated. This results in a 3D position and the 3D orientation of the drone (i.e., the attached cameras) for each recorded image pair. See \cite{Bimber2019IEEECGA} for more details on the AOS process.

For synthetic aperture visualization, a synthetic focal plane (SFP) has to be defined. With respect to the synthetic aperture plane (i.e., the drone's sampling plane, SAP), we parameterize the SFP with the relative plane distance $d$ and plane tilt orientation ($\theta,\phi$, in polar coordinates). As illustrated in Fig. \ref{fig:Fig1}b, a novel image can be computed for each SFP by integrating the pixel contributions of each single recording that project to the same point on the SFP. Under certain sampling conditions, it becomes statistically likely that unoccluded views of the same point on the SFP dominate over occluded views in these integrals. Consequently, objects above and below the focal plane (e.g., occluders) are suppressed while objects on the focal plane are amplified and become more visible. See \cite{Kurmi2018} and \cite{Kurmi2019Sensor}  for more details on the image integration process and the statistical principles and  of AOS. 

The complexity of synthetic aperture visualization increases linearly with the number of sample images to be integrated for a chosen SFP. A critical limitation of this, thus far, lies in finding the optimal SFP parameters ($d,\theta,\phi$) that reveal a widely focused and occlusion-free image of people on the ground, as their presence and location are unknown. Wrong SFP parameters lead to defocused integral images in which occlusion dominates. Exploring this parameter space manually through interactive visualization or by brute force search requires many tens to hundreds of thousands  sampling attempts, which is time consuming and error prone -- even if the parameter space is bound.  

In the following sections, we discuss and validate objective functions and optimization strategies that are suited for fast automatic visibility optimization in a thermal synthetic aperture visualization context. In contrast to manual sampling, we now find an optimal solution automatically and within a minimum number of iterations. 
%------------------------------------------------------------------
%---------------------------------------------------
\begin{figure*}[t!]
    \centering
    \includegraphics[width=\linewidth]{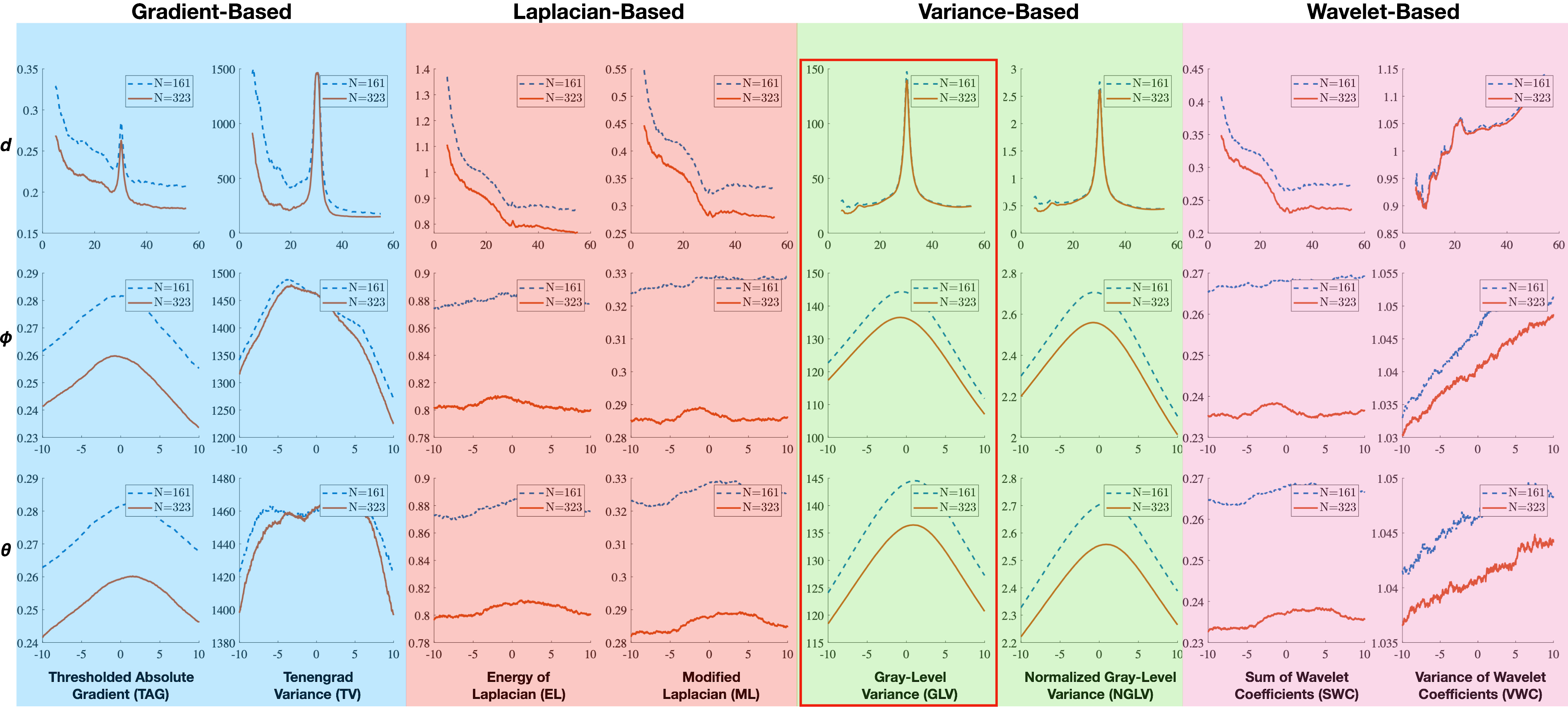}
    \caption{Comparison of fast (left) and robust (right) image focus metrices in different domains (gradient, Laplacian, variance, and wavelet). The conifer forest scan (Fig. \ref{fig:Fig3}, left column) was used for computing these plots. Plots for $\theta,\phi$ were calculated for the optimal $d$. The $y$ axes depicts the metric value, the $x$ axis is the distance from the SAP (top row) and angle offset from the SAP (center and bottom rows). Two different SA sampling densities are plotted: $N=323,N=161$ images.}
    \label{fig:Fig4}
    \vspace{-1.0em}
\end{figure*}
%---------------------------------------------------

\section{Objective Function for Visibility in Thermal Integral Images}
For parameter optimization, an adequate objective function is required that is capable of detecting improvement and degradation in visibility (i.e., focus and occlusion) for different SFP settings. Various focus metrics \cite{HUANG2007,Pertuz2013} exist that estimate the amount of defocus in conventional gray scale images based on gradients \cite{Chern2001,schlag1983}, Laplacians \cite{Lee1995,Nayar1994}, variance \cite{Firestone1991,sun2004}, or wavelets \cite{Yang2003,Huang2005}. However, it remains unclear how much such metrics are affected by occlusion. The presence of more or less strong occlusion in addition to more or less strong defocus is the main difference between our integral images and conventional images. 

Below, we prove that a simple variance-based metric is an ideal objective function for optimizing visibility in thermal integral images. 
We rely on the visibility function $\visibility$ derived in \cite{Kurmi2019Sensor}, that models occlusion (appearing in every SA image sample) as Bernoulli random variable with the probability being the occlusion density $\p$ and the number of recorded SA image samples $\N$:
%------------------------------------------------------------------------------
\begin{equation} 
    \visibility = 1-\p^2 - \frac{\p(1-\p)}{\N}.
    \label{eq:visbility_simplified}
\end{equation}
%------------------------------------------------------------------------------
Eqn.~\ref{eq:visbility_simplified} is based on the mean squared error ($\MSE$) between the resulting integral image $X$ and a hypothetical occlusion free image $S$ of the target (i.e., people on ground surface in our case) on a given SFP:
%------------------------------------------------------------------------------
\begin{equation}
    \visibility = 1 - \MSE = 1 -  \Ex{{\big(\X - \img \big)}^{2}},
    \label{eq:visibility}
\end{equation}
%------------------------------------------------------------------------------
where $\E$ is the expectancy (i.e., the mean). It assumes, however, a uniform target signal $S$ (variance $\sigmaSign=0$ and mean of $\meanSign=0$) and binary occluders $\noise$ ($\sigmaOccl=0$ and $\meanOccl=1$).
%------------------------------------------------------------------------------
If we extend Eqn. \ref{eq:visbility_simplified} to consider non-uniform targets and occluders, we obtain (see appendix for derivation):
%------------------------------------------------------------------------------
\begin{equation}
\visibility = 1 - \big( \p^2 + \frac{\p(1-\p)}{\N} \big) \big(\sigmaSign + (\meanOccl - \meanSign)^2\big) - \frac{\p}{\N}  \sigmaOccl
\label{eq:visibility_extended}
\end{equation}
%------------------------------------------------------------------------------
For any SFP setting, $\p$, $\N$, $\meanSign$, and $\meanOccl$ remain constant. When assuming a uniform distribution of occluders also $\sigmaOccl$ remains constant for every SFP slice through the occlusion volume. Only the variance of the target signal $\sigmaSign$ changes. Thus, the change in visibility  that is contributed to SFP variations is invariant to occlusion and sampling, but is proportional to the target signal's change of variance:
%------------------------------------------------------------------------------
\begin{equation}
\visibility(d,\theta,\phi) \propto \sigmaSign (d,\theta,\phi).
\label{eq:visibility_extended_simplified}
\end{equation}
%------------------------------------------------------------------------------
%XXXX Note, that the factor $\big( \p^2 - \frac{\p(1-\p)}{\N} \big)$ in Eqn.~\ref{eq:visibility_extended} is the probability of occlusion, the term $(\meanOccl - \meanSign)^2$ reflects the contrast between occluders and target, and the term $\frac{\p}{\N} \sigmaOccl$ is a signal averaging term. %, that reduces the variance of occluders, thus making them less visible. XXXX
The average contrast between occluders and target is the term $(\meanOccl - \meanSign)^2$ in Eqn.~\ref{eq:visibility_extended}. A high  occluder-target contrast reduces visibility. 
The constant factor $\big( \p^2 - \frac{\p(1-\p)}{\N} \big)$ in Eqn.~\ref{eq:visibility_extended} affects the average contrast as well as the signal variance. It reflects the level of occlusion in the integral bound within $\p$ (maximum) and $\p^2$ (minimum) for $\N=1$ and $\N=\infty$, respectively. The term $\frac{\p}{\N} \sigmaOccl$ describes the uniformity of occluders in the integral, where a high $\N$ also improves visibility.
All of these components are constant after recording. Although they do influence the overall visibility, they are independent of SFP variations. Consequently, a simple image variance metric is suitable for detecting an improvement or degradation of visibility in thermal integral images for changing SFP parameters $(d,\theta,\phi)$.     
%------------------------------------------------------------------
\section{Optimization Strategy}
Numerical optimization strategies are either derivative-based or derivative-free. While derivative-based methods (i.e., gradient-driven such as Newtons method, sequential quadratic programming, gradient descent, etc.) require continuous and differentiable objective functions, derivative-free techniques (i.e., optimization strategies entirely based on function values such as iterative pattern search or heuristics algorithms) support discontinuous objective functions without analytical or computationally estimated derivatives.   

It turns out that, for thermal integral images, variance-based metrics are nonlinear, deterministic, fairly smooth, continuous, partially differentiable, constrained, and locally convex (cf. Fig. \ref{fig:Fig4}). Therefore, global optimization techniques can be used in absence of proper parameter constraints. But since regions of local convexity are also relatively large for variance-based metrics, efficient local iterative-based non-linear optimization techniques lead to a significant reduction of iterations for a constrained range and/or a proper initial guess in $d,\theta,\phi$.  
%------------------------------------------------------------------
\section{Validation}
For validation, we chose the integral images' \textit{gray-level variance (GLV)} \cite{Firestone1991} as objective function, which has a linear complexity with respect to image resolution. For visibility optimization in $d,\theta,$ and $\phi$ we chose \textit{scatter search with gradient-based local optimizer (SS)} \cite{Ugray2007} without initial constrains, and \textit{sequential quadratic programming (SQP)} \cite{Spellucci1998} if the parameter range can be bound or an initial guess is available. 

Fig. \ref{fig:Fig3} illustrates results for AOS scans of four different scenes: conifer forest, broadleaf forest, mixed forest, open field (no occlusion), day light and low light conditions, as well as for static and moving targets. In all cases, a global optimum was reached that could not be improved further by manual adjustments or by brute force search. Note, that the remaining defocus in final thermal integral images (lower row in Fig. \ref{fig:Fig3}) is due to slight errors in drone pose estimation. This cannot be compensated. The reason for some targets appearing brighter than others is regionally more and less occlusion.

\begin{figure}[t!]
    \centering
    \includegraphics[width=\linewidth]{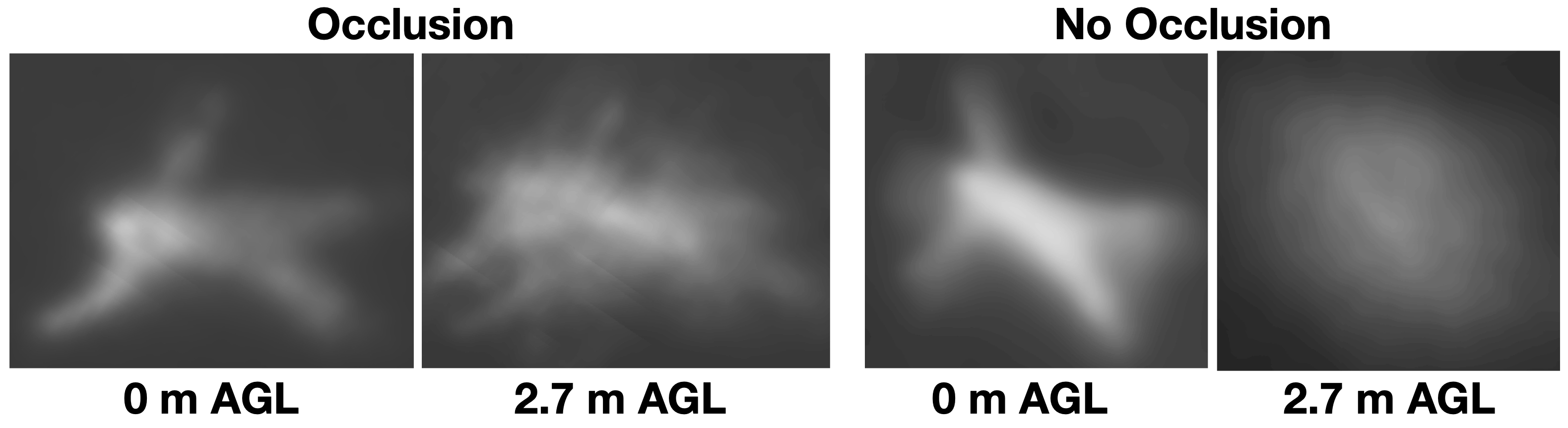}
    \caption{Focused and defocused thermal integral images for cases with occlusion (conifer forest in left column of Fig. \ref{fig:Fig3}) and without occlusion (open field in right column of Fig. \ref{fig:Fig3}). The numbers at the bottom indicate the distance of the SFP above ground level.}
    \label{fig:Fig2}
    \vspace{-1.5em} % table is more compact by adding vspaces <---- SPACE ADJUSTMENTS!
\end{figure}

Fig.~\ref{fig:Fig4} plots results of classical focus metrics in different domains (gradient, Laplacian, variance, wavelet) that are traditionally used for conventional images. We selected two representative examples (the fastest and the most robust, according to \cite{Pertuz2013,HUANG2007}) of each domain for comparison. As suggested by Eqn. \ref{eq:visibility_extended_simplified}, variance-based metrics lead to a clear and correct global optimum for thermal integral images. All other metrics, however, fail in general (see also supplementary video). This is very different to focus measures in conventional images, where Laplacian-based and wavelet-based metrics usually outperform variance-based metrics \cite{Pertuz2013,HUANG2007}. The reason for a diverse behavior with integral images lies in the effect of occluders that influences image gradients (first-order and second-order derivatives) and frequencies. As depicted in Fig.~\ref{fig:Fig2}, the point spread of a defocused target (i.e., its bokeh) is  structured by occluders while it is smooth without occluders. It is this structure that makes gradient-, Laplacian-, and wavelet-based metrics become variant to occlusion, while variance-based metrics remain occlusion-invariant (Eqn. \ref{eq:visibility_extended_simplified}). In all cases (with and without occlusion) the mean of the integral images is constant and is therefore independent of the SFP \cite{Kurmi2019Sensor}. The variance of the target, however, does change with respect to the SFP location.
%---------------------------------------------------
%---------------------------------------------------

%---------------------------------------------------
\begin{table}[b!]
\vspace{-1.5em} % table is more compact by adding vspaces <---- SPACE ADJUSTMENTS!
\begin{minipage}[t]{\linewidth}
\newcolumntype{Y}{>{\centering\arraybackslash}X}
\newcommand{\thd}[1]{\multicolumn{1}{Y}{\textbf{#1}}} 
\renewcommand\tabularxcolumn[1]{m{#1}}% for vertical centering text in X column
\renewcommand{\arraystretch}{1.3}
    \centering
    \caption{Number of optimization iterations for the scenes in Fig.~\ref{fig:Fig3} with SS and SQP using GLV as objective function.} \vspace{-1em} %<---- SPACE ADJUSTMENTS!
\begin{tabularx}{\linewidth}{r|ccccc} 
 \hline
 & \thd{Conifer} & \thd{Broad- leaf} & \thd{Mixed} & \thd{Low~Light, Motion} & \thd{Open Field}  \\ 
  \hline
 %\textbf{\textit{SS}} & 2119 & 3319 & 3466 & 1771 & 2766  \\ <-- old
 \textbf{\textit{SS}} & 2572 & 3434 & 3005 & 1468 & 3039  \\ 
 %\textbf{\textit{SQP}{$^*$}} & 90 & 110 & 119 & 93 & 75  \\ previously
 \textbf{\textit{SQP}{$^*$}} & 102 & 111 & 96 & 82 & 74  \\ 
 \hline
 \multicolumn{6}{c}{\footnotesize{$^*$ $d$ is restricted to 22 and 38 \textit{m} away from the SAP (approx. 30-35 \textit{m} AGL)}}
 \end{tabularx}
 \label{tab:Tab1}
\end{minipage}
\end{table}
%------------------------------------------------------------------
Tab. \ref{tab:Tab1} summarize performance measures of our approach. Note, that in our implementation on a standard PC, integrating 340 recordings in each iteration requires approx. 17-18 \textit{ms}. Thus, for unconstrained global optimizations approx. 30-60 \textit{s}, and for $d$-bound local optimization only approx. 1-2 \textit{s} are necessary.       

\section{Conclusion and Future Work}
In this article, we make two main contributions. We prove that the visibility of targets in thermal integral images is proportional to the variance of the targets' image -- which is invariant to occlusion. Therefore, efficient variance-based metrics are ideal objective functions for parameter optimization. While gradient-, Laplacian-, and wavelet-based metrics are normally preferred for focus measurements in conventional images, they usually fail for thermal integral images as they are effected by occlusion. With this finding, we describe and validate a fast and fully automatic parameter optimization for thermal synthetic aperture visualization. It replaces manual exploration of the parameter space, which is time consuming and error prone. 

Since our approach is also applicable to individual regions of interest, image-space partitioning can support optimizations of regions with locally planar sub-SFPs, instead of assuming one common SFP for the entire scene. Furthermore, since our objective function basically indicates mis-registration of thermal images after integration, it might be also applicable to guide registration methods that compensate the remaining defocus that is due to drone pose estimation errors. Both has to be investigated in the future. 

%------------------------------------------------------------------
\section*{Appendix}
In this appendix we discuss the statistical model of the mean squared error ($\MSE$) between an integral image $\X$ and an hypothetical occlusion free reference $\img$ (as used in Eqn. \ref{eq:visibility}):
\appendixMath{\begin{equation}
    \MSE = %\Ex{(\X-\img)^2} = %
    \Ex{{\big(\X - \img \big)}^{2}} %= %
    = \Ex{\X^2} - 2\Ex{\X \img} + \Ex{\img^2}   . %
    \label{eq:MSEdef} %
    %\Ex{(\X-\img)}^2 + \Var{(\X-\img)},
\end{equation}}
%
%
%where $\img$ and $\X$ are random variables. %, and $\Var(\X-\img)$ is the statistical expected variance. 
The integral image $\X$ is the average of $\N$ single image recordings, where each single image pixel is either occlusion free or occluded. %
We model this by the random variables $\img$ (occlusion free), $\noise$ (occluded), and $Z$ (determines if occluded or not). 
Thus, a pixel is $\p$ likely occluded ($Z_{i}=1$), thus $\noise_i$, or otherwise $(1-\p)$ likely occlusion free ($Z_{i}=0$), thus $\img$:
\appendixMath{\begin{align}
  \X = \frac{1}{\N} \sum_{i=1}^\N Z_{i}\noise_{i}  + (1 - Z_{i})\img .
\end{align}}
All random variables are i.i.d.\ with $Z_{i}$ %
following a Bernoulli distribution with success parameter $\p$ (i.e.,
$\Ex{Z_i}=\Ex{Z_i^2}=\p$; furthermore, note that $\E[Z_{i}(1-Z_{i} )] = 0$ is true). %
The random variable $\img$ follows a distribution with mean $\Ex{\img}=\meanSign$ and $\Ex{\img^2}=(\meanSign^2+\sigmaSign)$ and analogously $\noise_{i}$ follows a distribution with mean $\Ex{\noise_i}=\meanOccl$ and $\Ex{\noise_i^2}=(\meanOccl^2+\sigmaOccl)$. %

Thus, the first term of Eqn.~\ref{eq:MSEdef} expands to %
\newcommand{\OneZi}{(1\!-\!Z_{i})}%
\newcommand{\OneD}{(1\!-\!\p)}%
\appendixMath{\begin{equation}\begin{split}
%\label{eq:IntegralImagePower} 
\Ex{\X^2} &= \E\left[\left(\frac{1}{\N} \sum_{i=1}^\N Z_{i}\noise_{i}  + (1 - Z_{i})\R\right)^2\right]  \\
&= \frac{1}{\N^2} \E\left[\left(\sum_{i=1}^\N Z_{i}\noise_{i}  + (1 - Z_{i})\R\right)\left(\sum_{k=1}^\N Z_{k}\noise_{k}  + (1 - Z_{k})\R\right)\right] \text{.} % \\
\end{split}\end{equation}}
By applying the distributive law we get $\N$ terms with $i=k$ and $\N(\N-1)$ terms with $i\neq k$:
\appendixMath{\begin{equation}\begin{split}
\Ex{\X^2} &= \frac{1}{\N^2}\E\bigg[\sum_{i=1}^N \big( Z_{i}\noise_{i} + \OneZi \R \big)^2  \\ 
&\quad +  \sum_{i=1}^N\sum_{k\neq i}^{N}\big(Z_{i}\noise_{i}\!  + \! \OneZi\R\big) %
\big( Z_{k}\noise_{k} \! + \! \OneZi\R \big) \bigg] \text{,} %
\end{split}\end{equation}}
which further simplifies to %
\appendixMath{\begin{equation}\begin{split} \label{eq:EX2}
\Ex{\X^2}  &= %
\frac{1}{\N^2}\bigg(\N \big( \p(\sigmaOccl + \meanOccl^2) + \OneD (\sigmaSign +\meanSign^2)\big) %
\\ & \quad + {N(\N\!-\!1)} \big( \p^2\meanOccl^2 + 2\p \OneD \meanSign \meanOccl +\OneD^2(\sigmaSign + \meanSign^2) \big) \bigg) , \\
\end{split}\end{equation}}

The second term of Eqn.~\ref{eq:MSEdef} expands to %
\appendixMath{\begin{equation}\begin{split}
\label{eq:EXS}
\Ex{\X \img} &= %
\E\left[\frac{1}{\N} \sum_{i=1}^\N Z_{i}\noise_{i}\R  + \OneZi \R^2\right] \\
&= \p \meanSign \meanOccl  + \OneD (\sigmaSign + \meanSign^2) \text{.} \\%
\end{split}\end{equation}}%
%and %
%\begin{align*} \appendixMathFont 
%E[\R^2] &= \sigmaSign   + \meanSign^2 \text{.}
%\end{align*}
%

Using the expanded terms (Eqns.~\ref{eq:EX2} and \ref{eq:EXS}) in Eqn.~\ref{eq:MSEdef} yields
\appendixMath{\begin{align*}
\MSE %
&= %
\frac{1}{\N}\bigg( \p(\sigmaOccl + \meanOccl^2) + \OneD (\sigmaSign +\meanSign^2) %
\\ & \quad + {(\N-1)} \big( \p^2\meanOccl^2 + 2\p \OneD \meanSign \meanOccl +\OneD^2(\sigmaSign + \meanSign^2) \big) \bigg) \\
& \quad - 2 \left(  \p \meanSign \meanOccl  + \OneD (\sigmaSign + \meanSign^2) \right) %
+ \sigmaSign   + \meanSign^2 \\
&= \frac{1}{\N}\bigg(\OneD (\sigmaSign +\meanSign^2) + \p(\sigmaOccl + \meanOccl^2)\\&\quad - \big( \OneD^2(\sigmaSign + \meanSign^2) + 2\p\OneD\meanSign \meanOccl + \p^2\meanOccl^2 \big) \bigg) 
  \\&\quad + \OneD^2(\sigmaSign + \meanSign^2) - 2\OneD (\sigmaSign + \meanSign^2) + (\sigmaSign +\meanSign^2) %
\\&\quad + \p^2\meanOccl^2 + 2\p\OneD\meanSign \meanOccl - 2\p \meanSign \meanOccl%
\\&= \frac{1}{\N}\bigg(\p\OneD \sigmaSign  +\p\OneD (\meanSign^2 + \meanOccl^2 -2\meanSign \meanOccl) + \p\sigmaOccl\bigg)
  \\&\quad + \p^2\sigmaSign  + \p^2( \meanSign^2  + \meanOccl^2 -2\meanSign \meanOccl) \\
	&= \frac{\p\OneD}{\N} \sigmaSign  +\frac{\p\OneD}{\N} (\meanSign - \meanOccl)^2 + \frac{\p}{\N}\sigmaOccl 
  \\&\quad + \p^2\sigmaSign  + \p^2(\meanSign - \meanOccl)^2 \text{,}%
\end{align*}}
which finally simplifies to
\appendixMath{\begin{align}%
  %\MSE &= \bigg( \p^2 + \frac{\OneD\p}{N} \bigg) \big(\Var({\img}) + (\E({\noise}) - \E({\img}))^2\big) \\&+ \frac{\p}{\N}  \Var({\noise})
  \MSE = \left(\p^2 + \frac{\p\OneD}{\N}\right)  \left(\sigmaSign + (\meanOccl - \meanSign)^2\right)+ \frac{\p}{\N} \sigmaOccl \text{.}
\end{align}}%
\vspace{-.8em}
%%------------------------------------------------------------------
%\section*{Acknowledgment}
%XXX
%------------------------------------------------------------------

\ifCLASSOPTIONcaptionsoff
  \newpage
\fi

% -----------------------------------------------------------------------
%\printbibliography
\bibliographystyle{ieeetr}
\bibliography{main}
% -----------------------------------------------------------------------
\end{document}